\documentclass[letterpaper]{article} 
\usepackage[]{aaai25}  
\usepackage{times}  
\usepackage{helvet}  
\usepackage{courier}  
\usepackage[hyphens]{url}  
\usepackage{graphicx} 
\usepackage{natbib}  
\usepackage{caption} 
\usepackage{algorithm}
\usepackage{algorithmic}
\usepackage{amsmath} 
\usepackage{newfloat}
\usepackage{multirow}
\usepackage{listings}
\DeclareCaptionStyle{ruled}{labelfont=normalfont,labelsep=colon,strut=off} 
\floatstyle{ruled}
\newfloat{listing}{tb}{lst}{}
\floatname{listing}{Listing}
\nocopyright 

\title{CognitionCapturer: Decoding Visual Stimuli From Human EEG Signal With Multimodal Information}
\author {
    Kaifan Zhang\textsuperscript{\rm 1},
    Lihuo He\textsuperscript{\rm 1*}
    Xin Jiang\textsuperscript{\rm 1},
    Wen Lu\textsuperscript{\rm 1}
    Di Wang\textsuperscript{\rm 1}
    Xinbo Gao\textsuperscript{\rm 1,2}
}
\affiliations {
    \textsuperscript{\rm 1}School of Electronic Engineering, Xidian University, Xi’an, China\\
    \textsuperscript{\rm 2}Chongqing University of Posts and Telecommunications, Chongqing, China\\
}

\begin{document}

\maketitle

\begin{abstract}

Electroencephalogram (EEG) signals have attracted significant attention from researchers due to their non-invasive nature and high temporal sensitivity in decoding visual stimuli. However, most recent studies have focused solely on the relationship between EEG and image data pairs, neglecting the valuable ``beyond-image-modality" information embedded in EEG signals. This results in the loss of critical multimodal information in EEG. To address this limitation, we propose CognitionCapturer, a unified framework that fully leverages multimodal data to represent EEG signals. Specifically, CognitionCapturer trains Modality Expert Encoders for each modality to extract cross-modal information from the EEG modality. Then, it introduces a diffusion prior to map the EEG embedding space to the CLIP embedding space, followed by using a pretrained generative model, the proposed framework can reconstruct visual stimuli with high semantic and structural fidelity. Notably, the framework does not require any fine-tuning of the generative models and can be extended to incorporate more modalities. Through extensive experiments, we demonstrate that CognitionCapturer outperforms state-of-the-art methods both qualitatively and quantitatively. Code: https://github.com/XiaoZhangYES/CognitionCapturer.

\end{abstract}

%

\section{Introduction}

Since its inception, a fundamental challenge in brain decoding is optimally expressing the meaningful information within brain signals. Reconstructing visual stimuli from brain signals is one of the interesting tasks with exciting application prospects. Initially, pioneering work using fMRI data \cite{kay2008identifying, miyawaki2008visual, naselaris2009bayesian} validated the possibility of reconstructing visual stimuli from fMRI data and successfully decoded simple textures and shapes. More recently, with the rapid development of deep learning methods, the use of deep learning models to decode fMRI brain signals has produced significant advancements \cite{ren2021reconstructing, Takagi_2023_CVPR, scotti2024reconstructing}.

\begin{figure}[ht]
  \centering
  \includegraphics[width=\columnwidth]{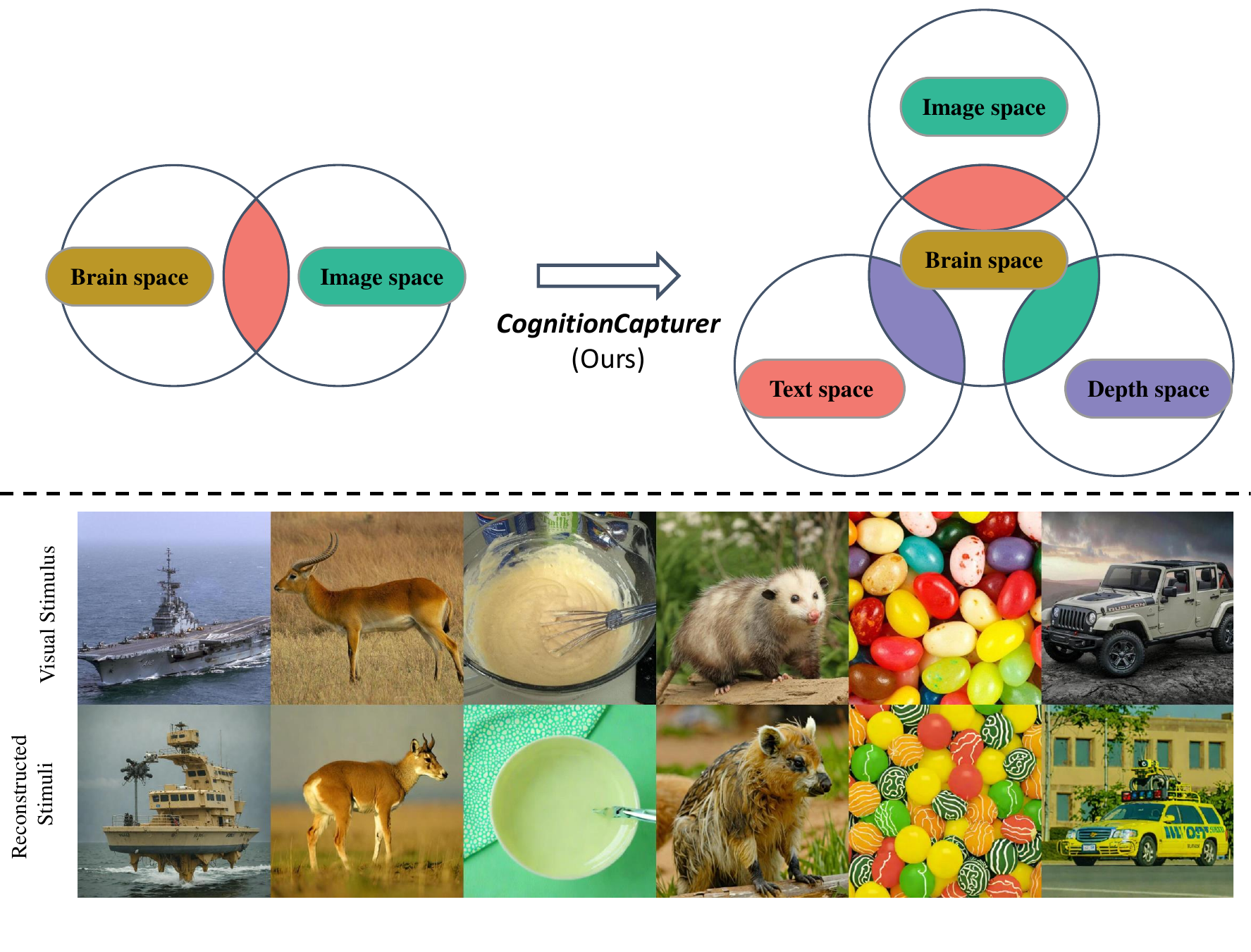}
  \caption{We believe that for image-EEG pairs, relying solely on the mutual information between images and EEG signals can lead to underutilization of EEG information. To address this issue, we utilize multimodal information to capture meaningful information in the EEG signals. The dashed lines in the figure below illustrate some of our successful reconstruction results.}
  \label{firstfig}
\end{figure}

However, brain signals exhibit diverse forms, among which EEG and MEG data offer high temporal resolution and portability, making them particularly suitable for real-time decoding compared to fMRI. This versatility has led to a broader range of downstream applications. Recent works \cite{benchetrit2024braindecodingrealtimereconstruction, song2024decoding, li2024visualdecodingreconstructioneeg} have attempted to align the brain-image modalities using EEG and MEG signals through contrastive learning. These approaches have achieved notable accuracy in decoding related visual stimuli.

However, the internal mechanisms of brain function are diverse and complex. Human perception of visual stimuli is influenced by both the characteristics of the visual stimuli and individual past experiences \cite{lupyan2020effects, du2023decoding}. Recent works \cite{benchetrit2024braindecodingrealtimereconstruction, song2024decoding, li2024visualdecodingreconstructioneeg} have primarily relied on the image modality as a reference for alignment, enabling the decoding of meaningful visual stimuli. Nonetheless, the objective of contrastive learning may lead to models that predominantly focus on the shared information between modalities, potentially overlooking the more diverse and complex ``beyond-image-modality'' information present in the brain signals.


To address this issue, we introduce a novel brain decoding model named CognitionCapturer, as illustrated in Fig. \ref{firstfig}. CognitionCapturer can be trained jointly with brain signals and multiple modalities, effectively capturing the shared information between brain signals and a broader spectrum of modalities.

Specifically, based on the understanding that brain data contains information ``beyond-image-modality'', we first extend image data using depth estimation models and image captioning models to construct a Image-Text-Depth multimodal aligned dataset. Then introduce Modality Expert Encoders, which focus on different EEG - single modality data. The embeddings obtained in this stage can be directly used for downstream tasks such as classification and retrieval. Subsequently, in the generation phase, we map the EEG embeddings to the CLIP image space via a diffusion prior and feed EEG embeddings associated with different modalities into a pre-trained image generation model, thus decoding fine-grained visual stimuli.

In contrast to previous methods, CognitionCapturer's training strategy enables models for different modalities to focus on capturing the relationships between information in EEG signals and modality-specific characteristics. This allows the model to capture fine-grained low-level visual information and abstract high-level semantic information. Furthermore, our proposed approach inherently possesses scalability, enabling the Modality Expert Encoder to be extended infinitely to any modality.

Another advantage of the proposed method is that the constructed dataset effectively decouples certain image features, allowing different Modality Expert Encoders to focus on structural and semantic features during training, thereby preventing fine-grained information from being overshadowed by coarse-grained information. The main contributions are as follows:

\subsection{Main Contribution}

\begin{itemize}
\item We propose CognitionCapturer, a contrastive learning-based model that effectively decodes brain signals from multiple modalities.
\item Using an alignment module and a pre-trained image generation model without any fine-tuning, we achieve fine-grained reconstruction of images with performance surpassing that of any single modality.
\item Through experiments, we validate the effectiveness and rationality of incorporating more modal information for brain signal decoding, providing new insights for subsequent research in neuroscience.
\end{itemize}

\begin{figure*}[ht]
\centering
\includegraphics[width=1\textwidth]{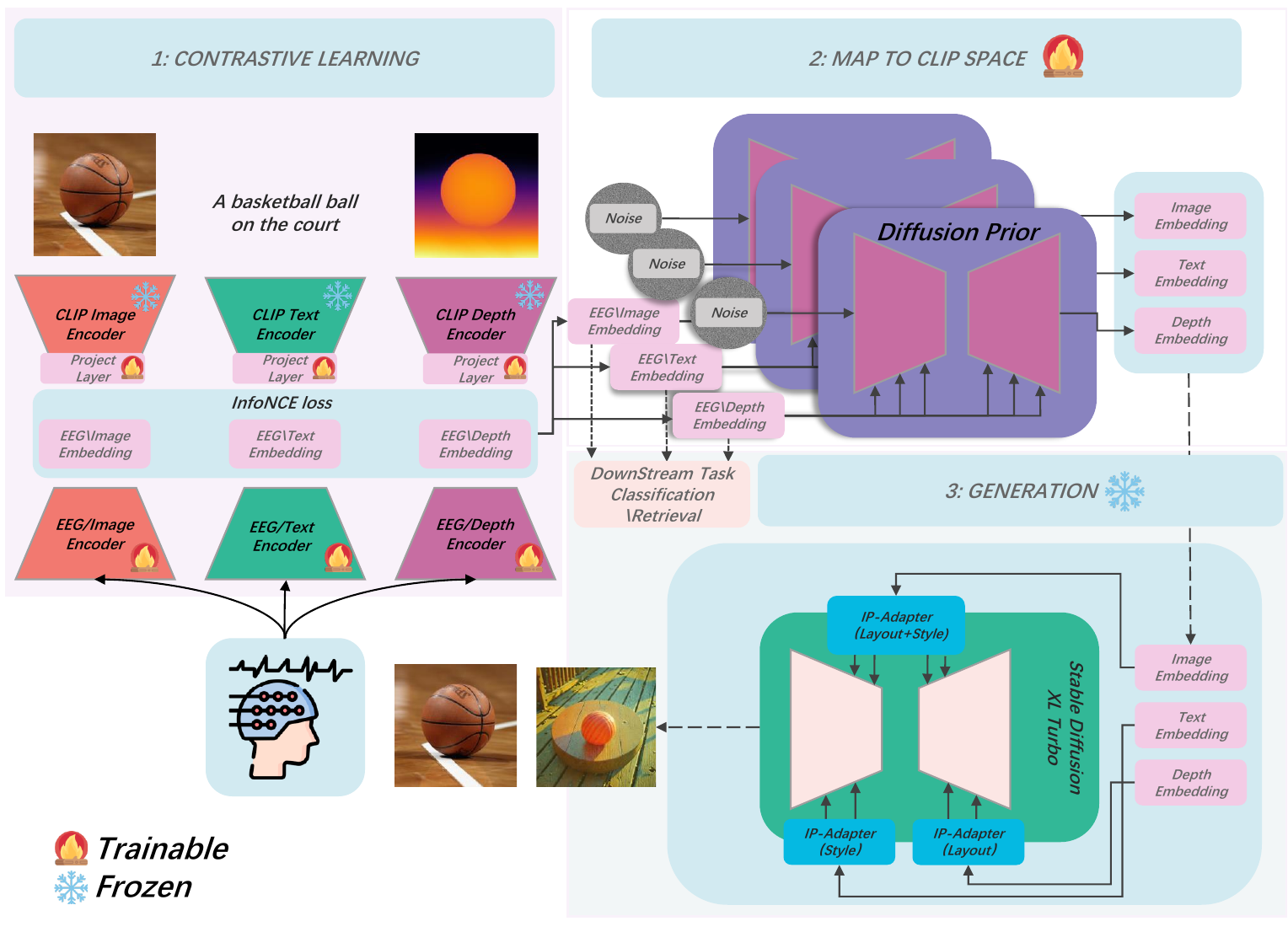} 
\caption{Overall framework of CognitionCapturer. 1: In the contrastive learning stage, different EEG-Modality data pairs are fed into different Modality Expert Encoders for processing. The embeddings obtained from the contrastive learning stage can be used for various downstream tasks. 2: 
To use pre-trained image generation models, we apply a Diffusion Prior model to map the EEG embeddings into CLIP space while retaining their original information. 3: Using pre-trained SDXL and IP-Adapters with different structures, we integrate the EEG embeddings from different modalities to reconstruct visual stimuli.}
\label{SturctureFig}
\end{figure*}

\section{Related Work}

\subsection{Decode Visual Stimuli from Brain Signal}

Decoding visual stimuli from fMRI brain signals has been widely studied and yielded successful results \cite{gu2024decoding, Takagi_2023_CVPR, scotti2024reconstructing, miyawaki2008visual, kay2008identifying}. However, the difficulty of acquiring fMRI data and its low temporal resolution pose challenges for practical applications. In contrast,  EEG signals offer higher temporal resolution and lower acquisition costs, leading researchers to attempt decoding visual stimuli from EEG. Early EEG decoding work typically relied on supervised learning methods and was limited to a finite set of image categories, overlooking the intrinsic relationship between visual stimuli and brain responses \cite{li2020perils, liu2023brain}
. Recently, \cite{song2024decoding, scotti2024reconstructing} successfully constructed an image decoding framework using a contrastive learning approach, achieving zero-shot recognition. \cite{li2024visualdecodingreconstructioneeg} built upon song's work \cite{song2024decoding} by further reconstructing decoded visual information into high-quality images using a diffusion model. However, these works only considered EEG-image modality pairs, neglecting the diversity of brain data. Compared with their approaches, our method successfully leverages multiple modalities of data to decode visual stimuli, resulting in superior performance.

\subsection{Contrastive Learning for Brain Decoding}

Contrastive learning, as an effective cross-modal learning approach, has achieved significant success in works such as CLIP, Moco, etc. \cite{radford2021learning, he2020momentum}, However, its effectiveness is closely related to the quality and scale of the data, and the selection of high-quality samples is crucial for improving model performance \cite{cherti2023reproducible}. Works that use contrastive learning to decode brain signals have also shown promising results. For instance, as a representative work, \cite{defossez2023decoding} utilizes a pre-trained speech encoder to decode speech from MEG signals through contrastive learning, and subsequently, \cite{benchetrit2024braindecodingrealtimereconstruction} adopts a similar idea to decode images from MEG.  A series of similar methods emerged subsequently \cite{song2024decoding, liu2023brainclipbridgingbrainvisuallinguistic, li2024visualdecodingreconstructioneeg}. However, during the process of using brain data for contrastive learning, the limited amount of brain signal data may lead the model to focus only on the most discriminative features. After transforming image modality into other modalities, since these modalities are less information-rich compared to image modality, this forces our model to attend to finer-grained features, thereby better representing EEG signals. 

\section{Method}

CognitionCapturer aims to address the loss of ``beyond-image-modality'' information in brain decoding. The method overview is depicted in Fig \ref{SturctureFig}, where EEG-Modality pairs\footnote{Specifically, the same EEG signals are divided into three pairs: EEG-Image, EEG-Text, and EEG-Depth. For consistency, we will refer to these collectively as EEG-Modality pairs. } are processed by dedicated Modality Expert Encoders to decouple the effective information from different modalities in the EEG signal. In our experiments, we observed that binding the brain modality with different modalities improves classification and reconstruction performance. Subsequently, through a diffusion prior, the EEG embedding space is mapped to the CLIP space and fed into assembled SDXL-turbo and IP-Adapters to reconstruct visual stimuli. 

\subsection{Modality Expert Encoder}    

CognitionCapturer uses modality pairs $(E, M)$, where $E$ represents the EEG signal and $M$ represents other modalities. For each modality pair $(E, M_{i})$, where $i$ represents the index of different modalities.

we construct a dedicated network $f_{i}$ and $g_{i}$, which we refer to as Modality Expert Encoders. This way, each modality pair $(E, M_{i})$ is mapped to the same dimension by its corresponding Modality Expert Encoder for subsequent constraints. 

In the encoding of the EEG data, raw EEG signals are typically represented as matrices $C \times T$, where $C$ denotes the number of electrode channels and $T$ denotes the number of time samples. Analysis of EEG signals primarily occurs along these two dimensions.

Our EEG encoder, based on a lightweight Transformer and STConv architecture \cite{li2024visualdecodingreconstructioneeg, vaswani2017attention}, effectively extracts topological and spatiotemporal information from EEG channels. The network structure is shown in Table \ref{model archi}. Specifically, we first process the raw EEG data  $e \in R^{ C \times T}$ through a layer of Transformer encoder and a linear transformation to organize the topological information, then feed it into a feature extraction module based on STConv to extract spatiotemporal features. Finally, a residual linear layer maps the features output by STConv to the same dimension as the target modality features. Detailed network descriptions are provided in the appendix. 

When extracting features for the target modality $ M_{i}$ paired with EEG data $E$, there are many successful pre-trained encoders that can effectively extract img, text, and depth features. Recent work \cite{zhang2022can} and our experiments indicate that CLIP image embeddings contain depth information. To be compatible with generative models and maintain distribution consistency initially, we used the Open CLIP ViT-H/14 \cite{radford2021learning} as both the visual and text encoder, and added a residual linear layer with the same dimension as the original features to ensure stability during training. 

\begin{table}[ht]
\centering
\begin{tabular}{lll}
\hline
Layer & Input Shape & Output Shape \\ \hline
Transformer Block & $(N, C, T)$& $(N, C, T)$ \\
Linear & $(N, C, T)$ & $(N, C, T)$ \\
STConv & $(N, C, T)$ & $(N, C_1, T_1)$ \\
Project Layer & $(N, C_1, T_1)$ & $(N, D)$ \\ \hline
\end{tabular}
\caption{Architecture of Modality Expert Encoder}
\label{model archi}
\end{table}

\subsection{Align EEG-Modality Pairs by Contrastive Learning}

After the modality pairs $(E, M_{i})$ are processed by their respective Modality Expert Encoder $f_{i}$ and $g_{i}$, they are encoded into the same dimension, resulting in embedding pairs $(e_{i}, m_{i})$. Here, $(e_{i}, m_{i})$ represents a set consisting of $n$ samples, i.e., $e_i = \{q^i_1, q^i_2, ..., q^i_n\},m_i = \{k^i_1, k^i_2, ..., k^i_n\}$.

Subsequently, for different $(e_{i}, m_{i})$ embedding pairs, we adopted an improved version of the infoNCE loss \cite{oord2019representationlearningcontrastivepredictive} as the loss function:


\begin{equation}
L_{E,M_{i}} = -\log\frac{L_{+}} 
{L_{+} + L_{-}}
\label{lossfunc}
\end{equation}

\begin{equation}
L_{+} = \sum_{P(idx) = 1}\exp(q^T_{idx} k_i / \tau)
\label{lossfunc+}
\end{equation}

\begin{equation}
L_{-} = \sum_{P(idx) = 0} \exp(q^T_{idx} k_i / \tau)
\label{lossfunc-}
\end{equation}

\begin{equation}
P(idx) =
\begin{cases}
1 & \text{when $idx$ is the same as image label} \\
0 & \text{otherwise}
\end{cases}
\end{equation}

In equation~\eqref{lossfunc+} and \eqref{lossfunc-}, $\tau$ is a scalar temperature parameter that controls the smoothness of the softmax distribution. Given that the same image is repeatedly viewed in EEG experiments \cite{gifford2022large}, multiple EEG data may correspond to the same image. This can create a contradictory phenomenon where the same data pairs are both pulled closer and pushed apart by the loss function. To address this, we utilize image index as supervisory information. Specifically, when $idx$ is the same in multiple EEG data, we choose to pull together all the EEG data and the corresponding modality data, thereby avoiding the contradictory phenomenon.In practice, we employ a symmetric loss $L_{E,M_{i}} + L_{M_{i},E}$. 

\subsection{Map EEG Embedding into CLIP Image Space}

After obtaining the aligned embeddings $e_{i}$ for EEG and $m_{i}$ for other modalities, due to the existence of the modality gap  and differences in distribution spaces \cite{scotti2024reconstructing}, directly using the EEG embedding $e_{i}$ would make it difficult for pre-trained generative models to identify effective information. 
Following the works of \cite{scotti2024reconstructing, li2024visualdecodingreconstructioneeg, ramesh2022hierarchicaltextconditionalimagegeneration}, we use a diffusion prior model to map the EEG embeddings $e_{i}$ to the CLIP space, thereby making the EEG embeddings recognizable by pre-trained generative models. In practice, we used the MSE loss to train our diffusion prior from scratch.

\begin{equation}
L_{\text{prior}} = {E}_{t \sim [1, T], m_i^{(t)} \sim q_t} \left[ ||f_\theta(m_i^{(t)}, t, e_i) - m_i||^2 \right]
\label{prior}
\end{equation}

In equation~\eqref{prior}, $m_i^{(t)}$ represents the CLIP embedding disturbed after a given diffusion timestep $t$, and $f_{\theta}$ denotes the diffusion prior network. The specific training details are provided in the Implementation Details section and supplementary material.

\begin{table*}[ht]
\centering
\setlength{\tabcolsep}{1mm}
\begin{tabular}{c|ccccccccccc}
\hline
Method & sub-01 & sub-02 & sub-03 & sub-04 & sub-05 & sub-06 & sub-07 & sub-08 & sub-09 & sub-10 & Ave \\ \hline
\multirow{2}{*}{\textit{CognitionCapturer (all)}}& \textit{31.41} & \textit{31.44} & \textit{38.19} & \textit{40.37} & \textit{24.44} & \textit{34.84} & \textit{34.65} & \textit{48.10} & \textit{37.42} & \textit{35.57} & \textit{35.64} \\
 & \textit{79.65} & \textit{77.80} & \textit{85.65} & \textit{85.80} & \textit{66.34} & \textit{78.75} & \textit{80.95} & \textit{88.60} & \textit{79.36} & \textit{79.29} & \textit{80.22} \\ \hline
\multirow{2}{*}{\textbf{CognitionCapturer (image)}} & \textbf{27.22} & \textbf{28.72} & \textbf{37.19} & \textbf{37.69} & \textbf{21.84} & \textbf{31.55} & \textbf{32.80} & \textbf{47.60} & \textbf{33.36} & \textbf{35.07} & \textbf{33.30} \\
 & 59.50 & \textbf{56.95} & \textbf{66.10} & \textbf{63.20} & \textbf{47.75} & \textbf{58.05} & 59.55 & \textbf{73.50} & \textbf{57.64} & \textbf{63.57} & \textbf{60.58} \\ \hline
\multirow{2}{*}{CognitionCapturer (text)} & 17.97 & 16.16 & 20.19 & 26.75 & 13.12 & 19.90 & 22.10 & 29.40 & 21.93 & 21.29 & 20.88 \\
 & 35.45 & 33.85 & 38.10 & 46.30 & 29.90 & 36.45 & 37.90 & 48.60 & 37.86 & 40.64 & 38.51 \\ \hline
\multirow{2}{*}{CognitionCapturer (depth)} & 23.10 & 21.85 & 29.65 & 34.40 & 15.75 & 27.50 & 30.90 & 36.90 & 27.14 & 26.86 & 27.41 \\
 & 57.40 & 53.25 & 61.65 & 65.50 & 40.25 & 50.20 & 54.55 & 60.20 & 49.00 & 49.14 & 54.11 \\ \hline
\multirow{2}{*}{BraVL \cite{du2023decoding}} & 6.1 & 4.9 & 5.6 & 5.0 & 4.0 & 6.0 & 6.5 & 8.8 & 4.3 & 7.0 & 5.8 \\
 & 17.9 & 14.9 & 17.4 & 15.1 & 13.4 & 18.2 & 20.4 & 23.7 & 14.0 & 19.7 & 17.5 \\ \hline
\multirow{2}{*}{NICE \cite{song2024decoding}} & 12.3 & 10.4 & 13.1 & 16.4 & 8.0 & 15.1 & 15.2 & 20.0 & 13.1 & 14.9 & 13.8 \\
 & 36.6 & 33.9 & 39.0 & 47.0 & 26.9 & 40.6 & 42.1 & 49.9 & 37.1 & 41.9 & 39.5 \\ \hline
\multirow{2}{*}{ATM \cite{li2024visualdecodingreconstructioneeg}} & 25.6 & 22.0 & 25.0 & 31.4 & 12.9 & 21.3 & 30.5 & 38.8 & 24.4 & 29.1 & 26.1 \\
 & \textbf{60.4} & 54.5 & 62.4 & 60.9 & 43.0 & 51.1 & \textbf{61.5} & 72.0 & 51.5 & 63.5 & 58.1 \\ \hline
\end{tabular}
\caption{Overall accuracy (acc\textpm std) of 200-way zero-shot classification: Top-1 and Top-5. The first line in each cell represents the Top-1 accuracy, and the second line represents the Top-5 accuracy. (In the calculation of CognitionCapturer (all)'s classification accuracy, if any Modality Expert Encoder correctly classifies a sample, the sample is considered correctly classified.) }
\label{top15results}
\end{table*}

\begin{table*}[ht]
\centering
\setlength{\tabcolsep}{1mm}
\begin{tabular}{l|lllllll}
\hline
 & \multicolumn{3}{c|}{Low-level} & \multicolumn{4}{c}{High-level} \\ \cline{2-8} 
Method (Averaged across subject) & PixCorr↑ & SSIM ↑ & AlexNet(2) ↑ & AlexNet(5) ↑ & Inception ↑ & CLIP ↑ & SwAV ↓ \\ \hline
CognitionCapturer (all) & 0.150 & 0.347 & 0.754 & 0.623 & 0.669 & 0.715 & 0.590 \\
CognitionCapturer (image) & 0.132 & 0.321 & 0.813 & 0.671 & 0.664 & 0.705 & 0.599 \\
CognitionCapturer (text) & 0.102 & 0.288 & 0.727 & 0.582 & 0.586 & 0.598 & 0.673 \\
CognitionCapturer (depth) & 0.104 & 0.370 & 0.796 & 0.638 & 0.565 & 0.579 & 0.686 \\
META-MEG \citeauthor{benchetrit2024braindecodingrealtimereconstruction}& 0.090 & 0.341 & 0.774 & 0.876 & 0.703 & 0.811 & 0.567 \\
MindEye-fMRI \citeauthor{scotti2024reconstructing}& 0.309& 0.323& 0.947& 0.978& 0.938& 0.941& 0.367\\ \hline
\end{tabular}
\caption{Quantitative comparison results on Things-EEG \cite{gifford2022large}  (compared to MEG data on Things-MEG \cite{hebart2023things} and fMRI data on NSD \cite{allen2022massive}). We report 7 different metrics to quantify the model's performance in reconstructing images at both low-level and high-level aspects.} 
\label{reconresults}
\end{table*}

\subsection{Generate visual stimulus with Multi-modal associated EEG embeddings }

After the EEG embeddings pass through the diffusion prior, they can be used like the original CLIP embeddings. Specifically, to reconstruct high-fidelity visual stimuli and effectively utilize information from three modalities, we employ Multi IP-Adapters \cite{ye2023ipadaptertextcompatibleimage} and SDXL-turbo \cite{sauer2023adversarialdiffusiondistillation} to simultaneously leverage embeddings from different modalities. As shown in Fig \ref{SturctureFig}'s generation phase, for the image modality, which contains the richest information, we use a full IP-Adapter to process the image embedding. For text and depth modalities, which focus on semantic and structural information respectively, we use modified versions of IP-Adapter, namely IP-Adapter-Style and IP-Adapter-Layout, to process the text and depth embeddings. This approach enables CognitionCapturer to reconstruct semantic information while preserving underlying visual details. 

\section{Experimental Setup}

\subsection{Datasets and Preprocessing}

We utilized Thing-EEG Dataset for our experiments. The Thing-EEG dataset \cite{gifford2022large} contains EEG data collected from 10 subjects under an RSVP paradigm. The training set comprises 1654 concepts, each associated with 10 images presented four times, resulting in a total of 66,160 EEG recordings. The test set includes 200 unique concepts, each represented by a single image repeated 80 times, totaling 16,000 EEG recordings. Both the training and test images are presented in a pseudorandom order to minimize habituation effects. Each image is displayed for 100 milliseconds followed by a blank screen for another 100 milliseconds to reduce blink-related and other artifacts. The raw EEG data were filtered between 0.1 Hz and 100 Hz, sampled at 1,000 Hz, and recorded using 63 channels. 

For EEG preprocessing, we follow the methodology outlined in \cite{song2024decoding, li2024visualdecodingreconstructioneeg}. We segment the EEG data into trials ranging from 0 to 1000 ms post-stimulus onset and perform baseline correction using the average value over the 200 ms period preceding the stimulus. All electrodes are retained, and the data are downsampled to 250 Hz. Multivariate noise normalization is applied to the training data, and the EEG repetitions for each image in the test set are averaged to improve the signal-to-noise ratio. Subsequently, to obtain a multimodally aligned dataset, we use BLIP2 \cite{li2023blip} for textual descriptions of the images and DepthAnything \cite{yang2024depth} for depth estimation, resulting in an aligned text and depth dataset.

\subsection{Implementation Details}

We implemented our method on a single GeForce RTX 2080 Ti GPU. following the training strategy described in \cite{song2024decoding}. The model was evaluated on the test set at the end of each epoch, with both training and testing conducted on separate subjects. For the training of the Modality Expert Encoder phase, we used the AdamW optimizer with a learning rate of 0.0003, a batch size of 1024, and trained for 20 epochs. Training for one subject took approximately 30 minutes.

Images were resized to 224×224 pixels and normalized before being processed by the Modality Expert Encoder. During the training of the diffusion prior, we used a batch size of 512, trained for 100 epochs, and set the number of inference steps to 50. The guidance scale was set to 7.5. In each batch, 10\% of the image embeddings were randomly replaced with noise. The embedding dimension was 1024.

In the generation process, we utilized SDXL-Turbo and IP-Adapter from Hugging Face. We set the inference steps for SDXL-Turbo to 5. When configuring the IP-Adapter, for the image modality, we used the full IP-Adapter with the scale set to 1. For the text and Depth modalities, we set the scale of their respective IP-Adapter's Layout block and Style block to 0, ensuring a focus on structural and semantic control in the reconstruction results.

 \begin{figure*}[ht]
  \centering
  \includegraphics[width=1\textwidth]{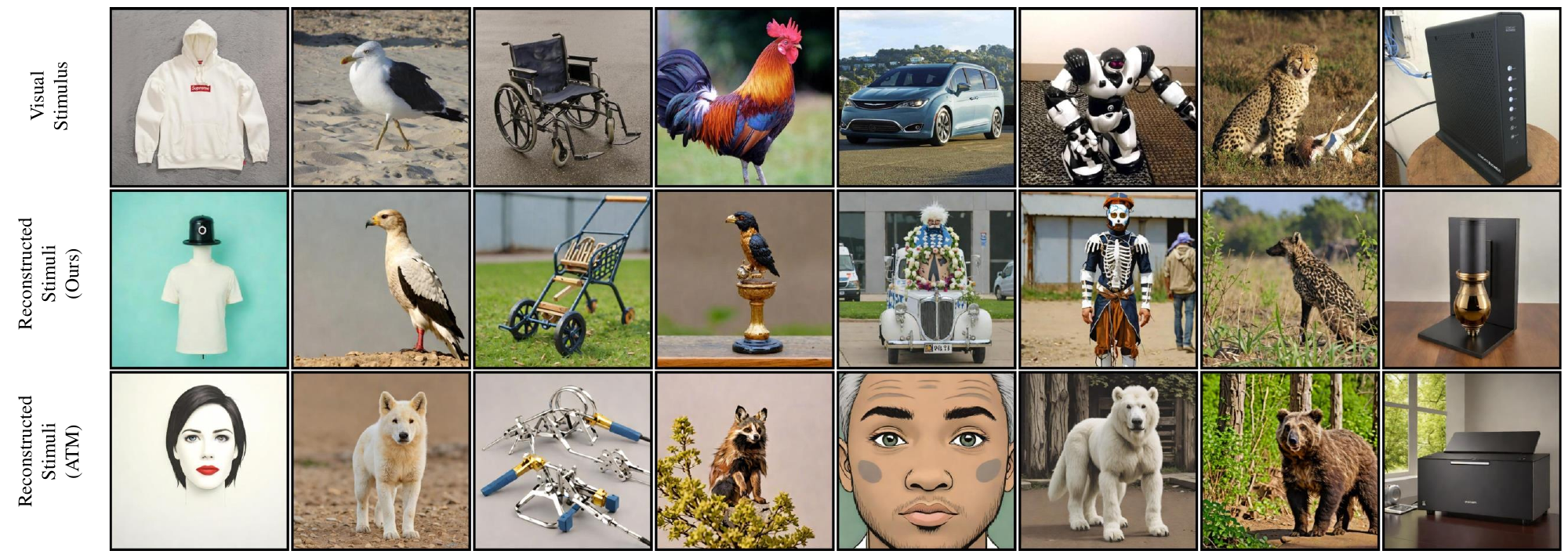}
  \caption{Visual Comparison. Selected reconstruction results from subject-08 show that our reconstructed visual stimuli exhibit finer-grained features.}
  \label{figcompare}
\end{figure*}

\section{Results and Discussion}

\subsection{Classification Performance}

The classification results of CognitionCapturer are shown in Table \ref{top15results}. We evaluated CognitionCapturer's ability to decode EEG embeddings based on different baseline modalities. To verify whether CognitionCapturer extracts complementary information across multiple modalities, we combined the top-5 results from three modalities, as shown in the upper bound row of Table \ref{top15results}. The results indicate that compared to previous work \cite{li2024visualdecodingreconstructioneeg, du2023decoding}, CognitionCapturer achieves state-of-the-art performance on the image modality. With the introduction of the text and depth modalities, the model gains access to more complementary information\footnote{Note: This does not represent the actual accuracy that can be achieved in practice but rather serves to demonstrate the effectiveness of the complementary information.}, leading to a significant increase in the potential amount of effective information. This suggests that complementary information across different modalities is indeed effective.

\subsection{Visual Stimuli Reconstruction Performance\footnote{More results can be found in the supplementary material.}}

Since subject-08 showed the highest classification results in both our model and ATM, we chose subject-08 for the comparison. Some of the visual stimuli reconstructed by CognitionCapturer are shown in Fig \ref{figcompare}. 

\begin{figure}[ht]
  \centering
  \includegraphics[width=\columnwidth]{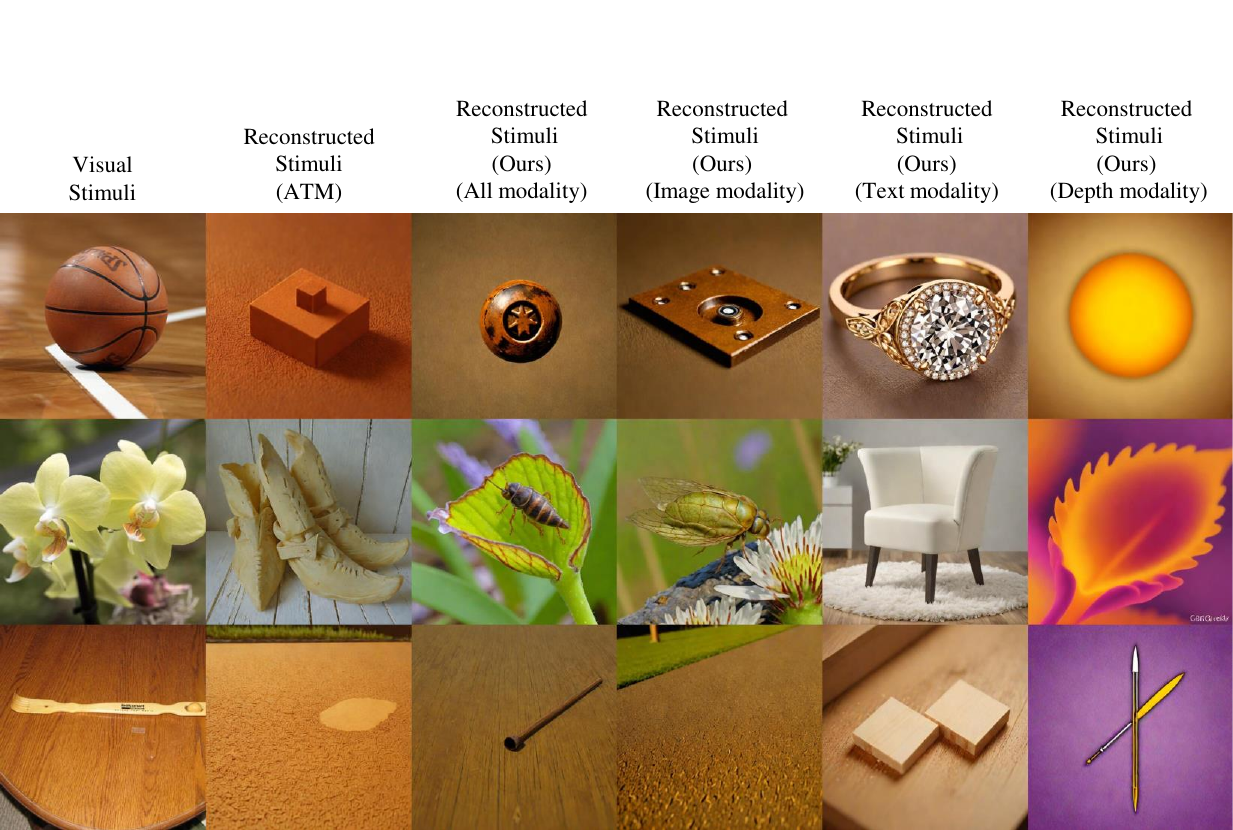}
  \caption{Reconstruction results of CognitionCapturer on different modality and comparison with prior work.}
  \label{figdepth}
\end{figure}

The results show that CognitionCapturer outperforms previous work \cite{li2024visualdecodingreconstructioneeg} in the fine-grained alignment of reconstructed visual stimuli. To further qualitatively analyze the effectiveness of CognitionCapturer's reconstruction, we recovered visual stimuli for each individual modality and compared them with the complete CognitionCapturer. As shown in Fig. \ref{figdepth}, there are differences in reconstruction performance when using single modalities: stimuli recovered only using the Text modality tend to be more abstract, while the Depth modality can better reconstruct structural information but performs poorly on semantic information. Notably, the image modality, which contains the richest information, sometimes loses certain details in its reconstructions. However, with the assistance of the Text and Depth modalities, CognitionCapturer recovers more reasonable visual stimuli. For instance, in Fig. \ref{figdepth}, when the visual stimulus is a basketball, the image modality misses the ``circular'' feature, whereas the depth modality retains this information well. 

To quantitatively compare our approach with the current state-of-the-art methods, we follow the evaluation metrics outlined in \cite{benchetrit2024braindecodingrealtimereconstruction} and conduct further quantitative comparisons on the reconstructed images. The results in Table \ref{reconresults} show that CognitionCapturer, when using all modality information, outperforms the use of a single modality in both low-level and high-level metrics. In low-level metrics, CognitionCapturer even matches or surpasses work using higher spatial resolution MEG signals. However, in high-level metrics, there remains a significant gap relative to MEG and fMRI signals, indicating that MEG and fMRI signals are easier to decode for meaningful information than EEG signals.

\subsection{How Different Modality Expert Encoders Focus on Brain Regions}

In the previous section, we analyzed the reconstruction results of CognitionCapturer. To provide evidence for the feasibility and interpretability of CognitionCapturer, we use Grad-CAM \cite{selvaraju2017grad} to visualize the regions of interest for different modality encoders. To mitigate the influence of individual subjects, we conducted an average analysis of the Grad-CAM results across all subjects' models. As shown in Fig. \ref{figtopo}(A), the raw EEG signal is heavily influenced by frontal lobe responses, whereas our Modality Expert Encoder primarily focuses on the occipital and temporal lobes, areas responsible for processing visual information \cite{dicarlo2007untangling}. Notably, compared to the Image Expert Encoder, which mainly attends to the occipital region, the Text Expert Encoder and Depth Expert Encoder attend to broader regions including both the occipital and temporal lobes. 

Surprisingly, the Depth Expert Encoder exhibits more significant attention to the right inferior temporal lobe, an area primarily involved in object recognition but less sensitive to object shape, size, and orientation \cite{epstein1998cortical}. We believe this is because depth information lacks many lower-level visual features such as color and texture, leaving only shape and depth information. Similar to the phenomenon of sensory compensation \cite{rauschecker1995compensatory}, this forces the model to seek higher-level brain information to ensure effective recognition of similar objects. This demonstrates that our modality-specific expert models reasonably focus on different brain regions, aligning with existing neuroscience theories.

\subsection{How Different Brain Area Interact with Visual Stimuli}

The analysis in the previous section demonstrated exciting results. To provide additional evidence for the effective interaction between EEG and image information, we further used Grad-CAM to visualize the image regions attended to by the embeddings produced by our Modality Expert Encoders and compared them with the original CLIP embeddings. 

As shown in Fig. \ref{figtopo}(B), first, in the original CLIP model, the text embedding focuses more on the object itself, while the image and depth embeddings have broader attention areas. Our Modality Expert Encoders yield EEG embeddings for different modalities that show similar results to those of CLIP. Specifically, the EEG embedding from the Text Expert Encoder focuses more on high-level information in the image, such as the baseball bats. In contrast, the Image and Depth Expert Encoders have broader attention over the image. Correspondingly, the brain regions attended to by the Image and Depth models are also more extensive compared to Text. This provides strong evidence for the interpretability of CognitionCapturer. 

\begin{figure}[ht]
  \centering
  \includegraphics[width=\columnwidth]{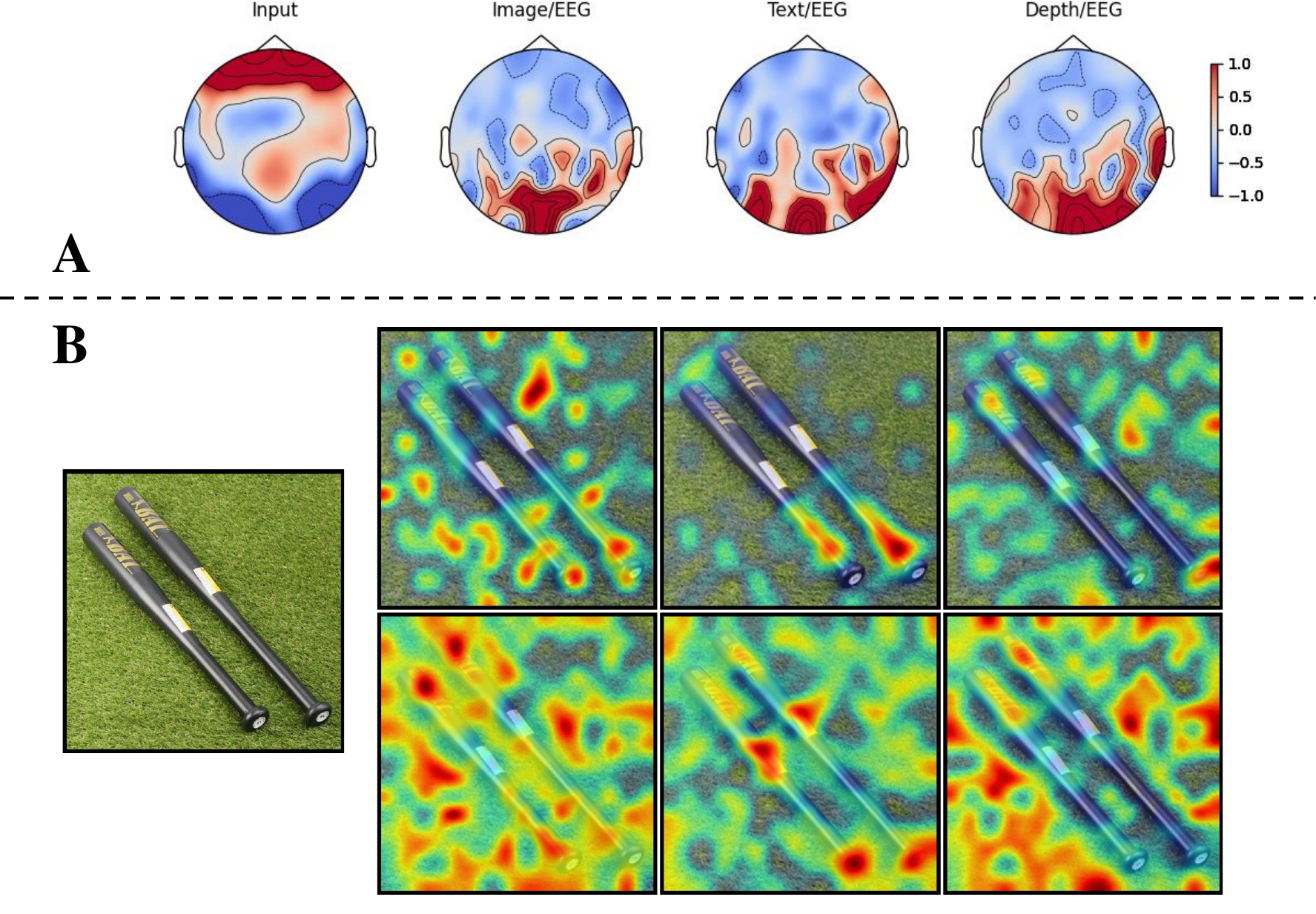}
  \caption{(A) The Grad-CAM results from different Modality Expert Encoders show the activation in the occipital and temporal lobes related to the input EEG signals. (B) The Grad-CAM results from different modality Expert Encoders on the brain signals corresponding to the example image, visualizing the regions of attention in the images and comparing them with the original CLIP embeddings.}
  \label{figtopo}
\end{figure}

\section{Conclusion}

In this work, we propose CognitionCapturer to extract multimodal representations from EEG signals and decode visual stimuli from them. Specifically, we introduce multiple Modality Expert Encoders to specialize in aligning EEG embeddings with those of different modalities, enabling the model to capture both semantic and structural information simultaneously. The analysis of brain activity and the interpretability of our model demonstrate that it successfully obtains meaningful representations of brain signals. This provides new insights for subsequent work in brain decoding. 

\section{Acknowledgments}

Thank you, Dr. Jili Xia, for your kindness and your advice on this work!

\begingroup
\small
\bibliography{aaai25}
\endgroup

\newpage

\section{Supplementary Material of CognitionCapturer: Decoding Visual Stimuli From Human EEG Signal With Multimodal Information}

\section{Details of CognitionCapturer}

\subsection{Modality Expert Encoder (EEG signal)}

The Modality Expert Encoder for EEG-Modality pairs uses a consistent structure adapted from \cite{li2024visualdecodingreconstructioneeg}. Specifically, the input EEG data is processed through a single-head attention model. Positional encoding is applied before passing the data through a transformer encoder, resulting in a vector of the same shape as the EEG input. The output is then passed through a linear layer. The resulting features are fed into a Temporal-Spatial Convolution (TSConv) module to generate the EEG embedding. Within the TSConv module, two consecutive convolution layers, a pooling layer, BatchNorm2d, and ELU are utilized for feature extraction.

\begin{itemize}
\item The first convolution layer extracts local temporal features, with an output channel count of 40, a kernel size of (1, 25), and a stride of (1, 1).
\item The second convolution layer extracts global channel-wise features, maintaining an output channel count of 40, a kernel size of (63, 1), and a stride of (1, 1).
\end{itemize}

Finally, the features are projected via a projection to generate the output. In the projection layer, the dimensions are first mapped to 1024 using a linear transformation. This is followed by a residual connection with an internal structure consisting of GELU and another linear layer. The output is then normalized using LayerNorm before being returned as the final output. For detailed dimension changes and parameters, refer to Table \ref{dimparam}.

\subsection{Modality Expert Encoder (Image Text Depth)}

The encoders for images, text, and depth all use the Image Text Encoder from \verb|OpenCLIP-ViT-H/14|. We consider depth as a form of image but add a projection layer to ensure stability. The structure of the projection layer is the same as the Project Linear structure in the Modality Expert Encoder.

\subsection{Diffusion Prior}

The model architecture of the Diffusion Prior is adapted from \cite{li2024visualdecodingreconstructioneeg}, and classifier-free guidance [reference here] is utilized during inference. Detailed structures can be found in the code file \verb|Scripts/train_align/diffusion_prior.py|.

\subsection{Impact of Batch Size and Learning Rate}

To identify the optimal hyperparameters, we experimented with various batch sizes and learning rates. Given that Subject-08 demonstrated the most representative performance, we focused solely on this subject for hyperparameter tuning. For brevity, we report only the classification performance using all modalities, with the results presented in Table \ref{batchlearn}.

\section{Reconstruction Performance and Results}

To supplement the reconstruction performance reported in the main paper, we present the reconstruction performance for each subject and modalities in Table \ref{reconall} - \ref{recondepth}. As the reconstruction performance of ATM\cite{li2024visualdecodingreconstructioneeg} was not specified for a particular subject or averaged across subjects, a direct comparison cannot be made under the same conditions. Therefore, we provide the test results obtained for each subject in the supplementary material. 

In the sample images, to avoid \textbf{cherry-picked} results and overestimating the capabilities of CognitionCapture, we follow the \cite{benchetrit2024braindecodingrealtimereconstruction}'s approach by ranking the reconstruction results according to the SwAV and PixCorr metrics from highest to lowest. We present the representative visual stimuli generated under the best, average, and worst metric conditions. We only display the results of the best-performing subject-08 and the worst-performing subject-05. The results are shown in Fig. 1 - 4. 

\section{More Model Visualization Results}

In this section, we present the visualization results obtained using Grad-CAM\cite{selvaraju2017grad} on the test set for each subject. These visualizations show the attention regions of the Modality Expert Encoders compared to the input attention regions of the respective subjects. The results are depicted in Fig. \ref{figtopo}. 

Regarding the Grad-CAM visualization of attention regions on images, we present the visualization results of the attention areas focused on by different Modality Expert Encoders for subject-08. The example images selected for visualization are the same as those used in the reconstruction results; see Fig. \ref{figvis}  for these results.

\begin{table*}[ht]
\centering
\begin{tabular}{lllll}
\hline
Layer & Type & Input Shape & Output Shape & Parameters \\ \hline
Spatial attention block & Positional Encoding + Attention & (batch, 63, 250) & (batch, 63, 250) & 553K \\
Linear & Linear & (batch, 63, 250) & (batch, 63, 250) & 63K \\
TSConv & Convolution + MaxPooling + BatchNorm & (batch, 63, 250) & (batch, 36, 40) & 104K \\
Temporal Aggregation & Dimention Transform & (batch, 36, 40) & (batch, 1440) & 0 \\
Project Linear & Residual Linear & (batch, 1440) & (batch, 1440) & 2527K \\ \hline
All &  & (batch, 63, 250) & (batch, 1024) & 3247K \\ \hline
\end{tabular}
\caption{Dimension changes and parameter counts in the modules of the Modality Expert Encoder. }
\label{dimparam}
\end{table*}

\begin{table*}[ht]
\centering
\begin{tabular}{lllll}
\hline
\multicolumn{1}{c}{\multirow{2}{*}{Batchsize   / Learning rate}} & \multicolumn{1}{c}{\multirow{2}{*}{1.00E-04}} & \multicolumn{1}{c}{\multirow{2}{*}{3.00E-04}} & \multicolumn{1}{c}{\multirow{2}{*}{6.00E-04}} & \multicolumn{1}{c}{\multirow{2}{*}{1.00E-03}} \\
\multicolumn{1}{c}{} & \multicolumn{1}{c}{} & \multicolumn{1}{c}{} & \multicolumn{1}{c}{} & \multicolumn{1}{c}{} \\ \hline
32 & 0.505 & 0.445 & 0.435 & 0.410 \\
64 & 0.485 & 0.490 & 0.430 & 0.435 \\
128 & 0.470 & 0.455 & 0.480 & 0.410 \\
256 & 0.470 & 0.495 & 0.505 & 0.460 \\
512 & 0.450 & 0.450 & 0.480 & 0.420 \\
1024 & 0.480 & 0.520 & 0.500 & 0.470 \\ \hline
\end{tabular}
\caption{The classification performance of CognitionCapturer under different batch sizes and learning rates for sub-08.}
\label{batchlearn}
\end{table*}

\begin{figure*}[ht]
\centering
\includegraphics[width=1\textwidth]{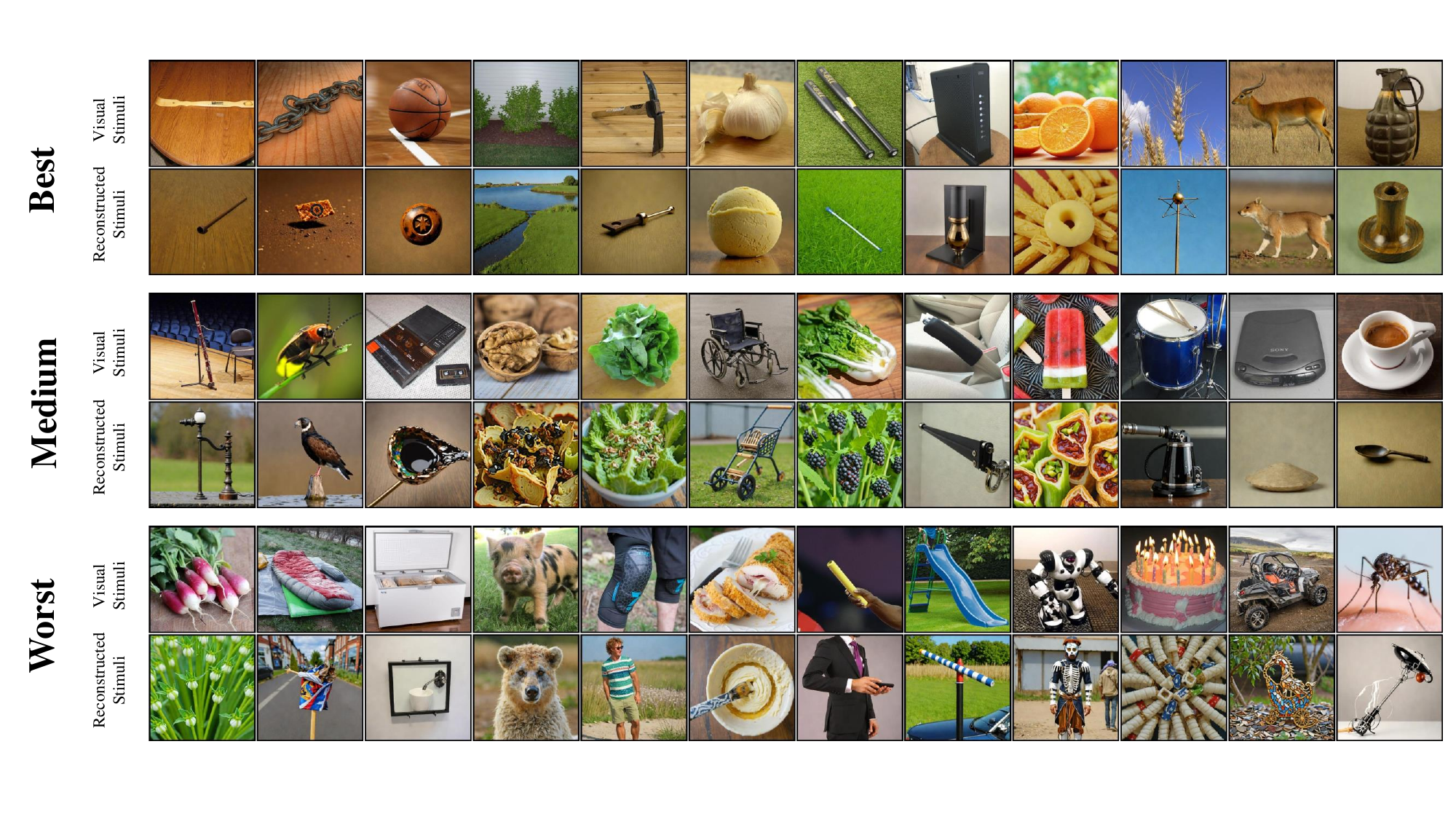} 
\caption{\textbf{Subject-08}'s Best, Medium, and Worst images selected based on the \textbf{Pixcorr} metric. }
\label{figpixcorr_sub08}
\end{figure*}

\begin{figure*}[ht]
\centering
\includegraphics[width=1\textwidth]{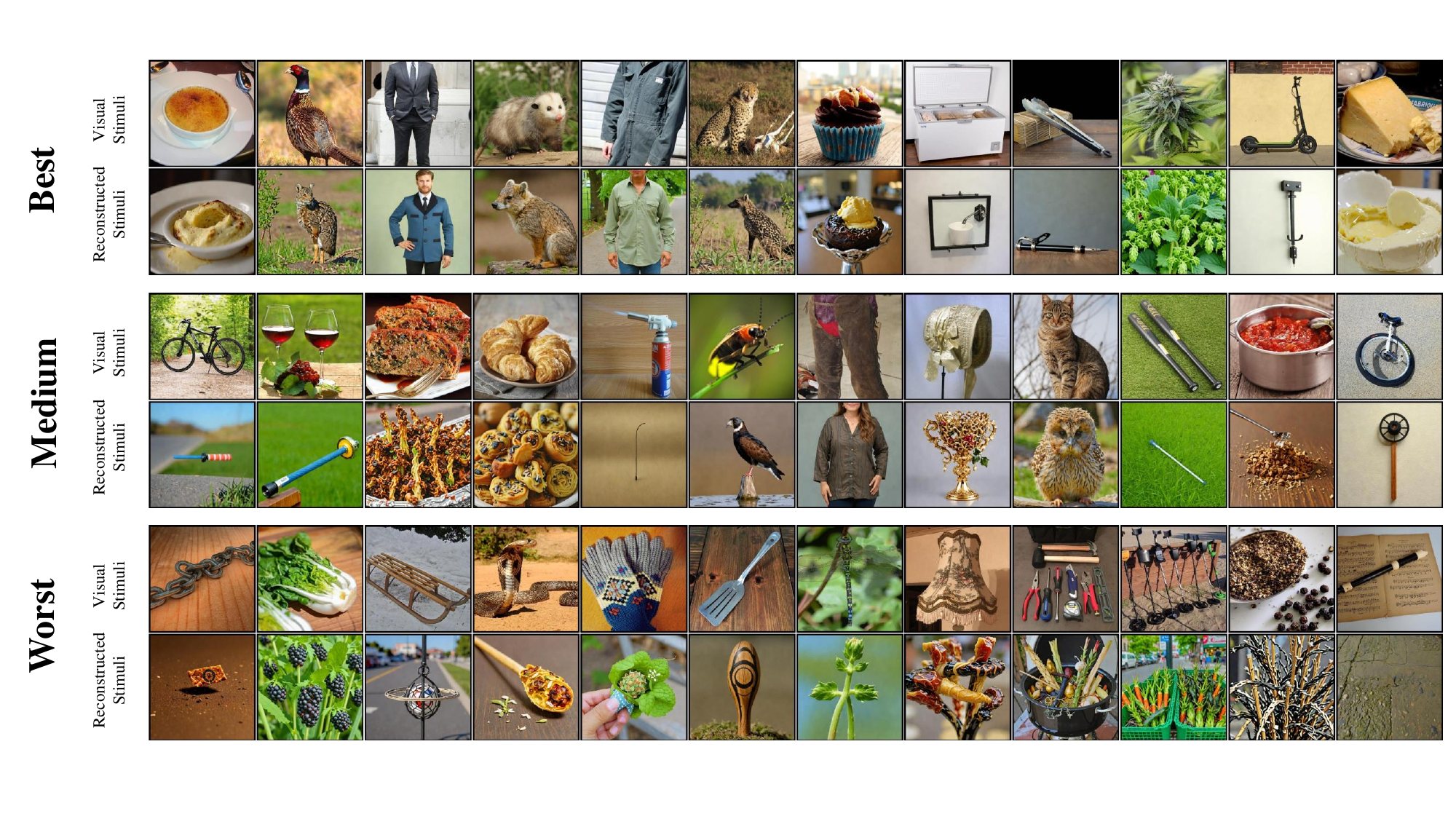} 
\caption{\textbf{Subject-08}'s Best, Medium, and Worst images selected based on the \textbf{SwAV} metric. }
\label{figswav_sub08}
\end{figure*}

\begin{figure*}[ht]
\centering
\includegraphics[width=1\textwidth]{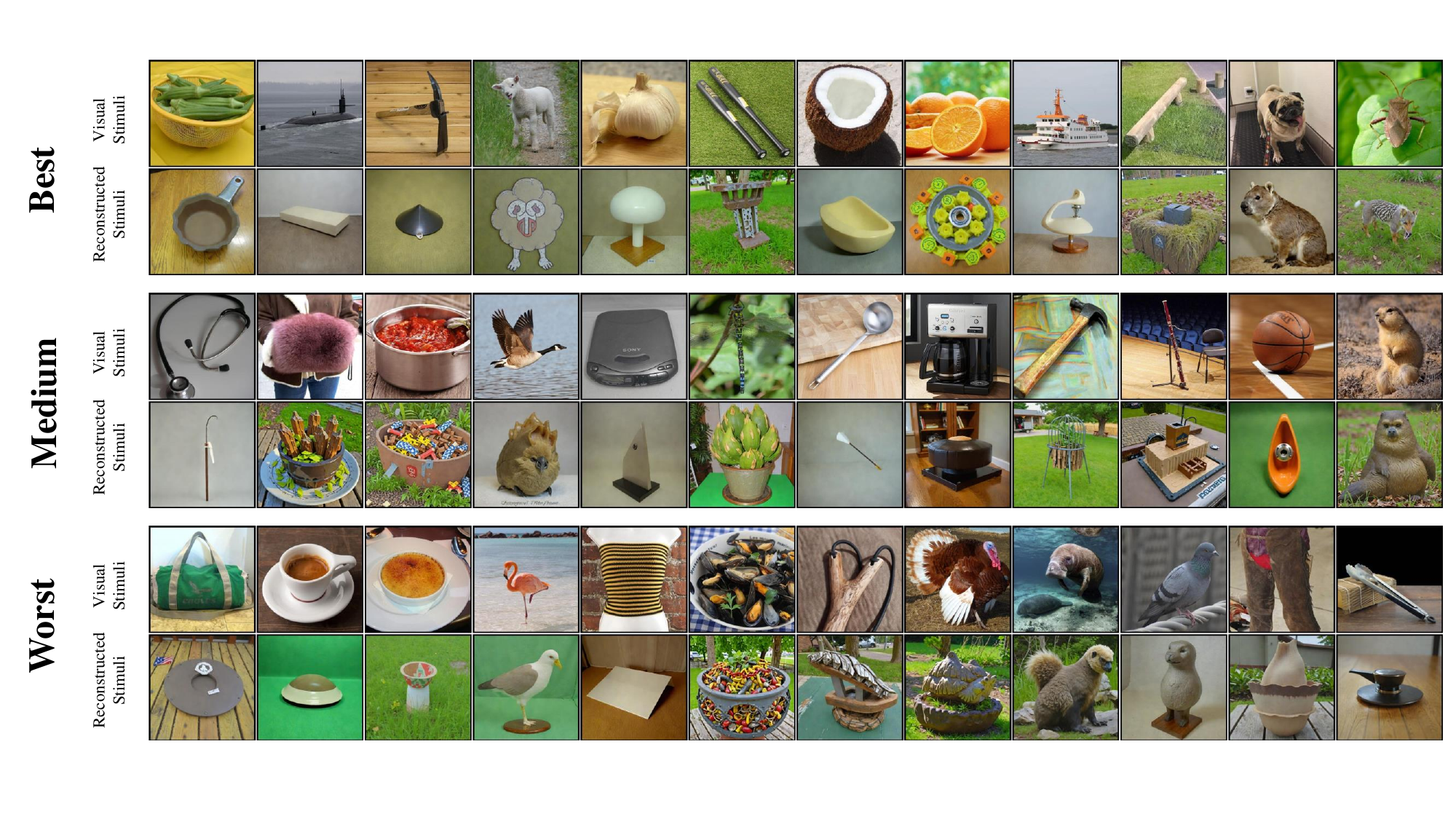} 
\caption{\textbf{Subject-05}'s Best, Medium, and Worst images selected based on the \textbf{Pixcorr} metric. }
\label{figpixcorr_sub05}
\end{figure*}

\begin{figure*}[ht]
\centering
\includegraphics[width=1\textwidth]{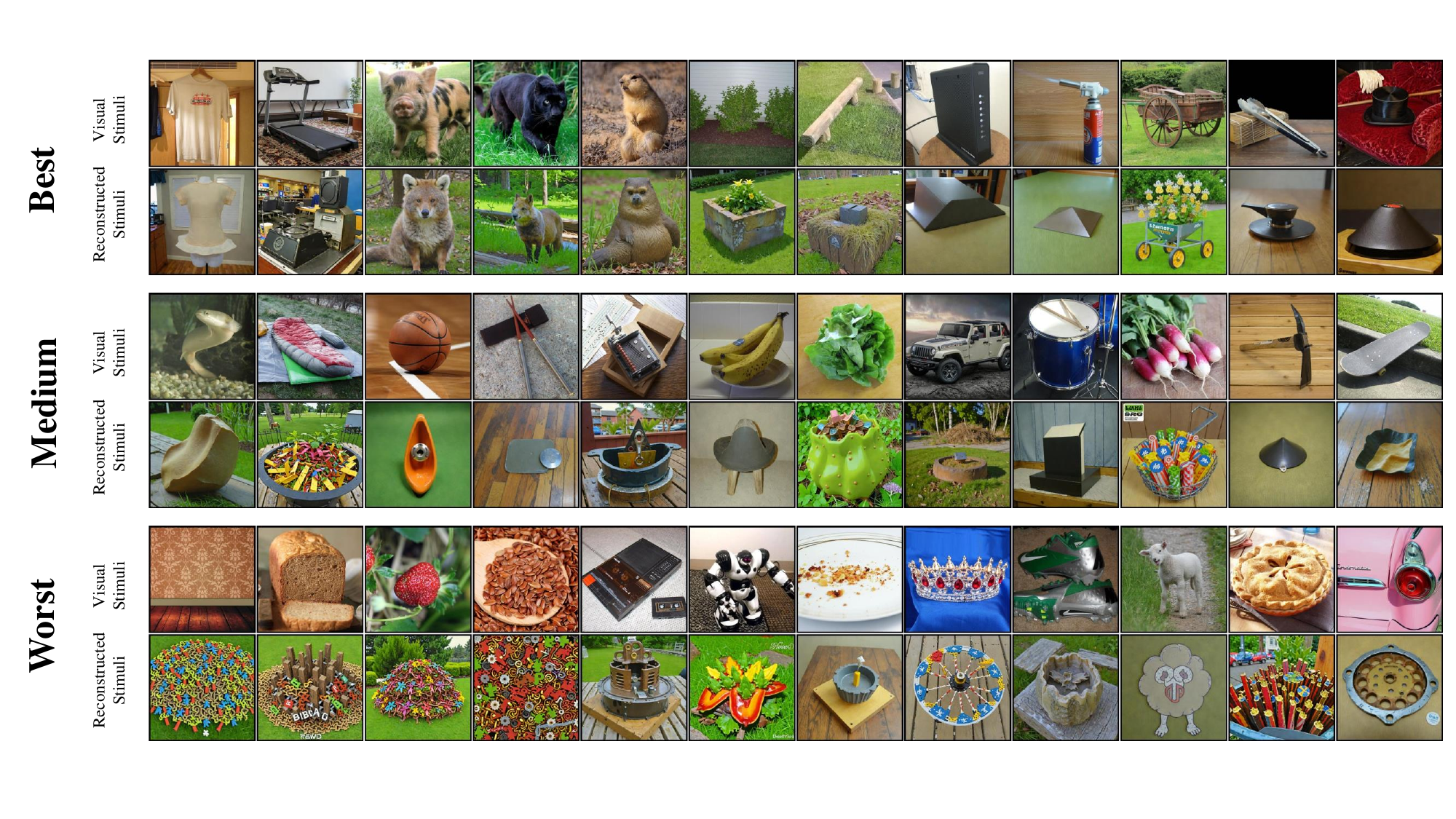} 
\caption{\textbf{Subject-05}'s Best, Medium, and Worst images selected based on the \textbf{SwAV} metric. }
\label{figswav_sub05}
\end{figure*}

\begin{table*}[ht]
\centering
\begin{tabular}{llllllll}
\hline
 & \multicolumn{3}{c|}{Low-level} & \multicolumn{4}{c}{High-level} \\ \cline{2-8} 
Subject & Pixcorr↑ & SSIM↑ & AlexNet(2) ↑ & AlexNet(5) ↑ & Inception↑ & CLIP↑ & SwAV↓ \\ \hline
1 & 0.148 & 0.334 & 0.741 & 0.626 & 0.666 & 0.711 & 0.592 \\
2 & 0.147 & 0.344 & 0.764 & 0.618 & 0.661 & 0.725 & 0.590 \\
3 & 0.140 & 0.307 & 0.715 & 0.549 & 0.690 & 0.710 & 0.603 \\
4 & 0.166 & 0.355 & 0.801 & 0.660 & 0.701 & 0.765 & 0.543 \\
5 & 0.130 & 0.343 & 0.731 & 0.639 & 0.594 & 0.655 & 0.611 \\
6 & 0.152 & 0.337 & 0.748 & 0.620 & 0.646 & 0.688 & 0.630 \\
7 & 0.145 & 0.355 & 0.777 & 0.623 & 0.731 & 0.721 & 0.576 \\
8 & 0.175 & 0.366 & 0.760 & 0.610 & 0.721 & 0.744 & 0.577 \\
9 & 0.148 & 0.337 & 0.731 & 0.623 & 0.625 & 0.692 & 0.605 \\
10 & 0.152 & 0.389 & 0.773 & 0.664 & 0.657 & 0.736 & 0.569 \\
Ave & 0.150 & 0.347 & 0.754 & 0.623 & 0.669 & 0.715 & 0.590 \\ \hline
ATM & / & 0.345 & 0.776 & 0.866 & 0.734 & 0.786 & 0.582 \\ \hline
\end{tabular}
\caption{The reconstruction performance of CognitionCapture when using \textbf{ALL} modalities.}
\label{reconall}
\end{table*}

\begin{table*}[ht]
\centering
\begin{tabular}{llllllll}
\hline
 & \multicolumn{3}{c|}{Low-level} & \multicolumn{4}{c}{High-level} \\ \cline{2-8} 
Subject & Pixcorr↑ & SSIM↑ & AlexNet(2) ↑ & AlexNet(5) ↑ & Inception↑ & CLIP↑ & SwAV↓ \\ \hline
1 & 0.126 & 0.317 & 0.812 & 0.653 & 0.655 & 0.700 & 0.595 \\
2 & 0.109 & 0.309 & 0.809 & 0.638 & 0.634 & 0.702 & 0.602 \\
3 & 0.137 & 0.300 & 0.803 & 0.604 & 0.651 & 0.689 & 0.609 \\
4 & 0.125 & 0.328 & 0.853 & 0.732 & 0.738 & 0.781 & 0.551 \\
5 & 0.116 & 0.327 & 0.769 & 0.668 & 0.618 & 0.667 & 0.612 \\
6 & 0.138 & 0.280 & 0.787 & 0.665 & 0.592 & 0.643 & 0.654 \\
7 & 0.135 & 0.330 & 0.842 & 0.700 & 0.695 & 0.711 & 0.598 \\
8 & 0.154 & 0.327 & 0.830 & 0.655 & 0.711 & 0.748 & 0.583 \\
9 & 0.138 & 0.310 & 0.799 & 0.670 & 0.664 & 0.673 & 0.603 \\
10 & 0.138 & 0.378 & 0.825 & 0.725 & 0.676 & 0.736 & 0.580 \\
Ave & 0.132 & 0.321 & 0.813 & 0.671 & 0.664 & 0.705 & 0.599 \\ \hline
\end{tabular}
\caption{The reconstruction performance of CognitionCapture when using \textbf{IMAGE} modality.}
\label{tab:reconimg}
\end{table*}

\begin{table*}[ht]
\centering
\begin{tabular}{llllllll}
\hline
 & \multicolumn{3}{c|}{Low-level} & \multicolumn{4}{c}{High-level} \\ \cline{2-8} 
Subject & Pixcorr↑ & SSIM↑ & AlexNet(2) ↑ & AlexNet(5) ↑ & Inception↑ & CLIP↑ & SwAV↓ \\ \hline
1 & 0.114 & 0.309 & 0.722 & 0.551 & 0.568 & 0.591 & 0.678 \\
2 & 0.105 & 0.280 & 0.716 & 0.589 & 0.575 & 0.604 & 0.679 \\
3 & 0.104 & 0.214 & 0.700 & 0.551 & 0.557 & 0.553 & 0.730 \\
4 & 0.117 & 0.341 & 0.761 & 0.628 & 0.662 & 0.636 & 0.624 \\
5 & 0.105 & 0.303 & 0.680 & 0.546 & 0.540 & 0.607 & 0.681 \\
6 & 0.111 & 0.273 & 0.716 & 0.566 & 0.591 & 0.600 & 0.675 \\
7 & 0.098 & 0.278 & 0.731 & 0.594 & 0.608 & 0.611 & 0.661 \\
8 & 0.078 & 0.267 & 0.776 & 0.615 & 0.589 & 0.572 & 0.695 \\
9 & 0.080 & 0.306 & 0.701 & 0.578 & 0.550 & 0.585 & 0.659 \\
10 & 0.109 & 0.308 & 0.769 & 0.599 & 0.621 & 0.626 & 0.649 \\
Ave & 0.102 & 0.288 & 0.727 & 0.582 & 0.586 & 0.598 & 0.673 \\ \hline
\end{tabular}
\caption{The reconstruction performance of CognitionCapture when using \textbf{TEXT} modality.}
\label{tab:recontext}
\end{table*}

\begin{table*}[ht]
\centering
\begin{tabular}{llllllll}
\hline
 & \multicolumn{3}{c|}{Low-level} & \multicolumn{4}{c}{High-level} \\ \cline{2-8} 
Subject & Pixcorr↑ & SSIM↑ & AlexNet(2) ↑ & AlexNet(5) ↑ & Inception↑ & CLIP↑ & SwAV↓ \\ \hline
1 & 0.116 & 0.340 & 0.798 & 0.618 & 0.556 & 0.561 & 0.701 \\
2 & 0.106 & 0.368 & 0.789 & 0.633 & 0.559 & 0.601 & 0.668 \\
3 & 0.097 & 0.365 & 0.775 & 0.621 & 0.565 & 0.578 & 0.710 \\
4 & 0.093 & 0.359 & 0.843 & 0.694 & 0.587 & 0.625 & 0.670 \\
5 & 0.082 & 0.417 & 0.744 & 0.582 & 0.533 & 0.528 & 0.699 \\
6 & 0.107 & 0.385 & 0.767 & 0.568 & 0.521 & 0.539 & 0.692 \\
7 & 0.115 & 0.375 & 0.812 & 0.641 & 0.572 & 0.578 & 0.676 \\
8 & 0.081 & 0.370 & 0.852 & 0.652 & 0.586 & 0.586 & 0.671 \\
9 & 0.119 & 0.361 & 0.764 & 0.642 & 0.543 & 0.554 & 0.697 \\
10 & 0.128 & 0.363 & 0.818 & 0.732 & 0.627 & 0.641 & 0.670 \\
Ave & 0.104 & 0.370 & 0.796 & 0.638 & 0.565 & 0.579 & 0.686 \\ \hline
\end{tabular}
\caption{The reconstruction performance of CognitionCapture when using \textbf{DEPTH} modality.}
\label{recondepth}
\end{table*}

\begin{figure*}[ht]
\centering
\includegraphics[width=1\textwidth]{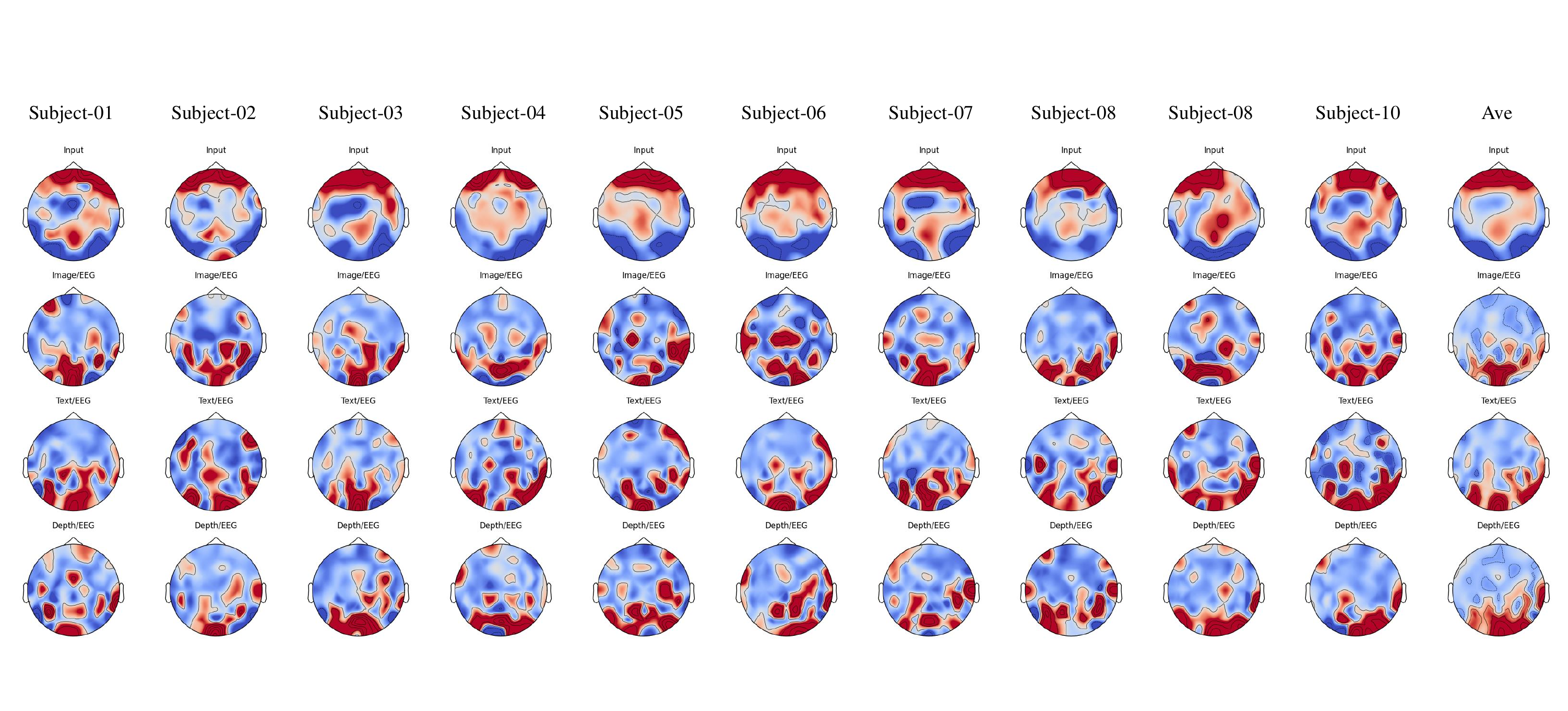} 
\caption{The input topographies of the EEG signals for all subjects, along with the brain regions attended to by the different Modality Expert Encoders.}
\label{figtopo}
\end{figure*}
\begin{figure*}[ht]
\centering
\includegraphics[width=1\textwidth]{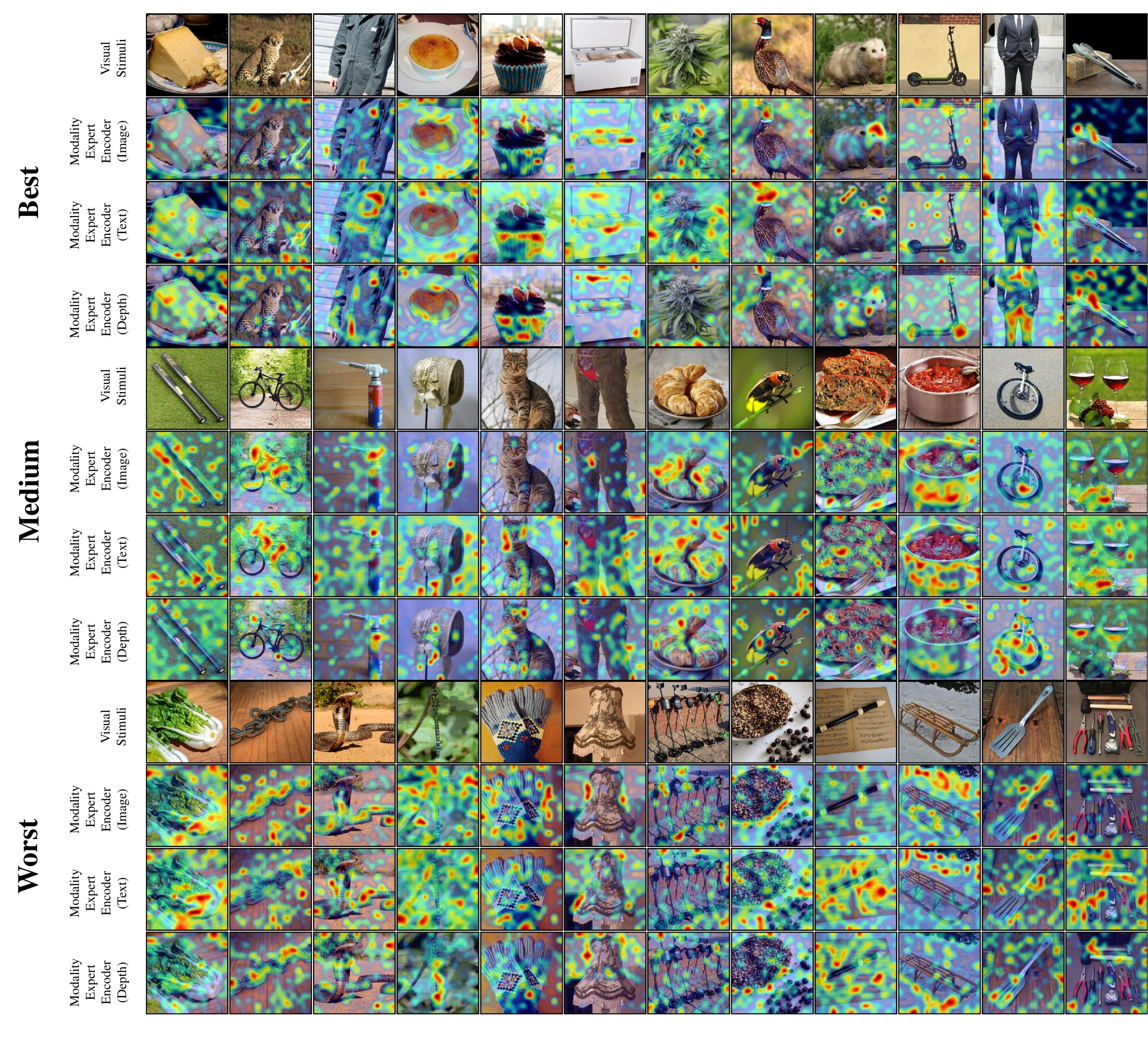} 
\caption{Grad-CAM visualizations of the image regions attended to by different Modality Expert Encoders on example images}
\label{figvis}
\end{figure*}

\end{document}